%% file: IFTOMM_Chablat_Caro_Angeles.tex
\documentclass[twocolumn,10pt]{IFTOMM}
\usepackage{psfrag}
\usepackage{epsfig}
\usepackage{array}
\usepackage{amssymb}
\usepackage{subeqn}
\usepackage{subfigure}

\input{macros.tex}

\begin{document}
\title{The Kinetostatic Optimization of a Novel Prismatic Drive}


\author{
    \begin{tabular}{c}
    D. Chablat\thanks{E-mail: Damien.Chablat@irccyn.ec-nantes.fr} \quad S. Caro\thanks{E-mail: Stephane.Caro@irccyn.ec-nantes.fr}\\
    Institut de Recherche en Communications et Cybern\'etique de Nantes,UMR CNRS n$^\circ$ 6597 \\
  1 rue de la No\"e, 44321 Nantes, France\\
    \end{tabular}
}

\maketitle

\input{01_abstract}
\input{02_Introduction}
\input{03_SynthesisPlanarCamMechanisms}
\input{04_DesignParamPressAngleHertzPress}
\input{05_Applications}
\input{06_Conclusions}
\input{07_Bibliography}

\end{document}

%% file: macros.tex
\newcommand{\be}{\begin{equation}}
\newcommand{\ee}{\end{equation}}
\newcommand{\beq}{\begin{equation}}
\newcommand{\eeq}{\end{equation}}
\newcommand{\bed}{\begin{displaymath}}
\newcommand{\eed}{\end{displaymath}}
\newcommand{\beqa}{\begin{eqnarray}}
\newcommand{\eeqa}{\end{eqnarray}}
\newcommand{\beqann}{\begin{eqnarray*}}
\newcommand{\eeqann}{\end{eqnarray*}}
\newcommand{\bseq}{\begin{subequations}}
\newcommand{\eseq}{\end{subequations}}

\newcommand{\ba}{\begin{array}}
\newcommand{\ea}{\end{array}}

\newcommand{\negr}[1]{{\bf {#1}}}


%% file: 01_abstract.tex
\begin{abstract}
The design of a mechanical transmission taking into account the
transmitted forces is reported in this paper. This transmission is
based on {\em Slide-o-Cam}, a cam mechanism with multiple rollers
mounted on a common translating follower. The design of Slide-o-Cam,
a transmission intended to produce a sliding motion from a turning
drive, or vice versa,  was reported elsewhere. This transmission
provides pure-rolling motion, thereby reducing the friction of
rack-and-pinions and linear drives. The pressure angle is a suitable
performance index for this transmission because it determines the
amount of force transmitted to the load vs.\ that transmitted to the
machine frame. To assess the transmission capability of the mechanism, 
the Hertz formula is
introduced to calculate the stresses on the rollers and on the cams.
The final transmission is intended to replace the current
ball-screws in the {\em Orthoglide}, a three-DOF parallel robot for
the production of translational motions, currently under development
for machining applications at {\em \'Ecole Centrale de Nantes}.
\end{abstract}

%% file: 02_Introduction.tex
\section{Introduction}
In robotics and mechatronics applications, whereby motion is
controlled using a piece of software, the conversion from rotational
motion to translational one is usually realized by means of {\em
ball-screws} or {\em linear actuators}. The both are gaining
popularity. However they present some drawbacks. On the one hand,
ball-screws comprise a high number of moving parts, their
performance depending on the number of balls rolling in the shaft
groove. Moreover, they have a low load-carrying capacity, due to the
punctual contact between the balls and the groove. On the other
hand, linear bearings are composed of roller-bearings to figure out
the previous issue, but these devices rely on a form of direct-drive
motor, which makes them expensive to produce and maintain.

\begin{figure}[htb]
 \begin{center}
   \psfrag{Roller}{Roller}
   \psfrag{Follower}{Follower}
   \psfrag{Conjugate cams}{Conjugate cams}
   \epsfig{file = 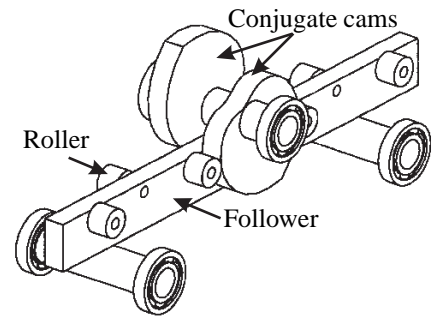,scale = 0.5}
   \caption{Layout of Slide-o-Cam}
   \label{fig001}
 \end{center}
\end{figure}

A novel transmission, called {\it Slide-o-Cam} is depicted in
Fig.~\ref{fig001} and was introduced in
\cite{Gonzalez-Palacios:2000} to transform a rotational motion to a
translational one. Slide-o-Cam is composed of four main elements:
($i$) the frame; ($ii$) the cam; ($iii$) the follower; and ($iv$)
the rollers. The input axis on which the cams are mounted, named
\emph{camshaft}, is driven at a constant angular velocity by means
of an actuator under computer-control. Power is transmitted to the
output, the translating follower, which is the roller-carrying
slider, by means of pure-rolling contact between the cams and the
rollers. The roller comprises two components, the pin and the
bearing. The bearing is mounted to one end of the pin, while the
other end is press-fit into the roller-carrying slider.
Consequently, the contact between the cams and rollers occurs at the
outer surface of the bearing. The mechanism uses two conjugate
cam-follower pairs, which alternately take over the motion
transmission to ensure a positive action; the rollers are thus
driven by the cams throughout a complete cycle. Therefore, the main
advantages of cam-follower mechanisms with respect to the other
transmissions, which transform rotation into translation are: ($i$)
the lower friction; ($ii$) the higher stiffness; and ($iii$) the
reduction of wear.

Many researchers have devoted their efforts to study contact stress
distribution and predict surface fatigue life in machine parts under
different types of loading. Indeed, when two bodies with curved
surfaces, for example, a cam and a roller, are pressed together, the
contact is not linear but a surface. The stress occurred may
generate failures such as cracks, pits, or flaking in the material.
Heinrich Rudolf Hertz (1857-1894) came up with a formula to evaluate
the amount of surface deformation when two surfaces (spherical,
cylindrical, or planar) are pressed each other under a certain force
and within their limit of elasticity.

%% file: 03_SynthesisPlanarCamMechanisms.tex
\section{Synthesis of Planar Cam Mechanisms}
\begin{figure}[htb]
 \begin{minipage}[b]{8cm}
 \begin{center}
   \psfrag{f}{$\bf f$}
   \psfrag{p}{$p$}    \psfrag{e}{$e$}    \psfrag{p}{$p$}    \psfrag{s}{$s$}
   \psfrag{d}{$d$}    \psfrag{x}{$x$}   \psfrag{y}{$y$}    \psfrag{P}{$P$}
   \psfrag{C}{$C$}    \psfrag{mu}{$\mu$}
   \psfrag{u}{$u$}       \psfrag{v}{$v$}
   \psfrag{delta}{$\delta$}
   \psfrag{b2}{$b_2$}   \psfrag{b3}{$b_3$}
   \psfrag{a4}{$a_4$}   \psfrag{psi}{$\psi$}
   \psfrag{theta}{$\theta$}
   \psfrag{O1}{$O_1$}   \psfrag{O2}{$O_2$}
   \centerline{\epsfig{file = 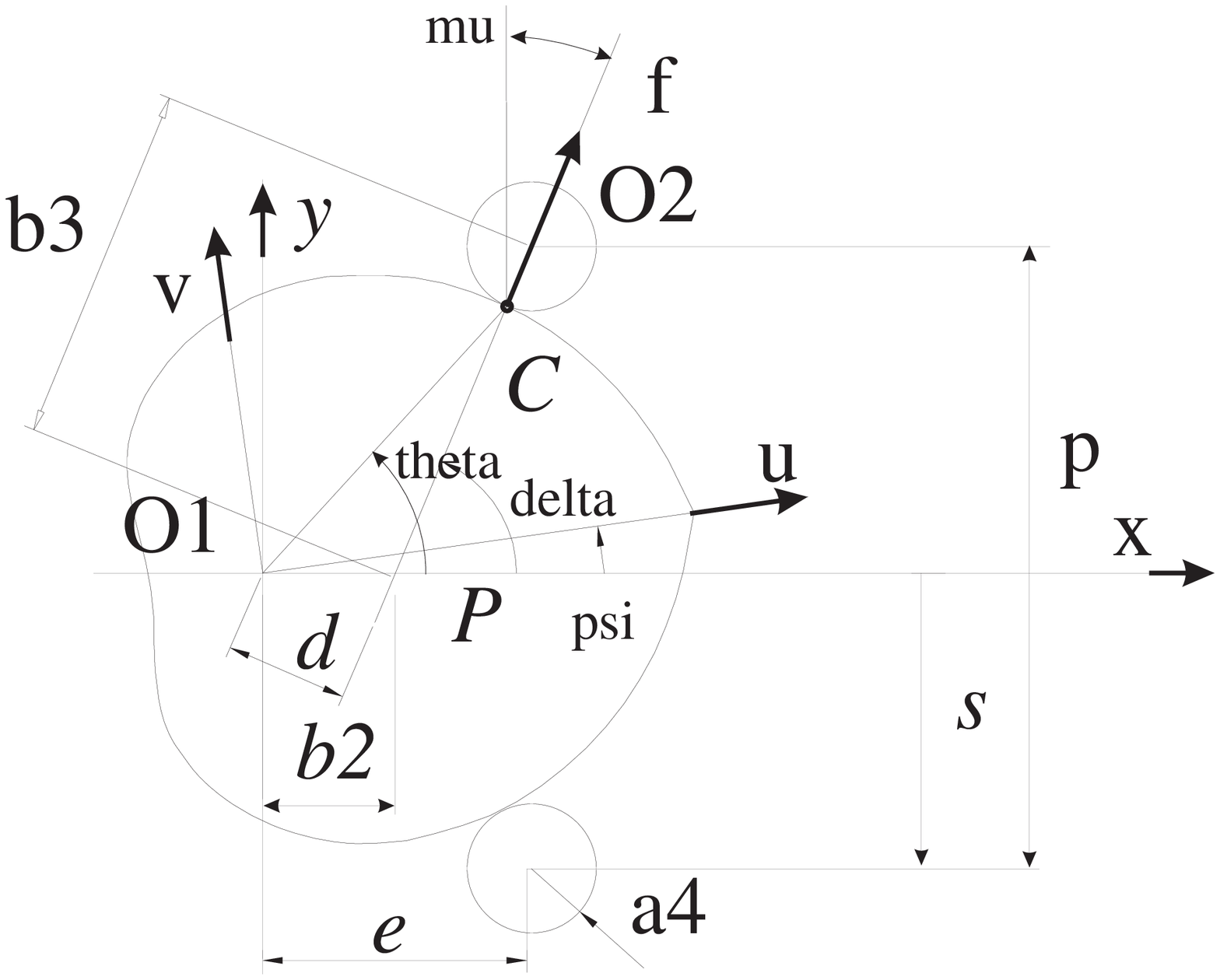,scale = 0.35}}
  \caption{Parameterization of \goodbreak Slide-o-Cam}
  \label{fig003}
 \end{center}
 \end{minipage}
 \begin{minipage}[b]{8cm}
 \begin{center}
   \psfrag{O1}{$O_1$}
   \psfrag{p}{$p$}    \psfrag{x}{$x$}    \psfrag{y}{$y$}
   \psfrag{u}{$u$}    \psfrag{v}{$v$}    \psfrag{x}{$x$}
   \psfrag{s(0)}{$s(0)$}
   \centerline{\epsfig{file = 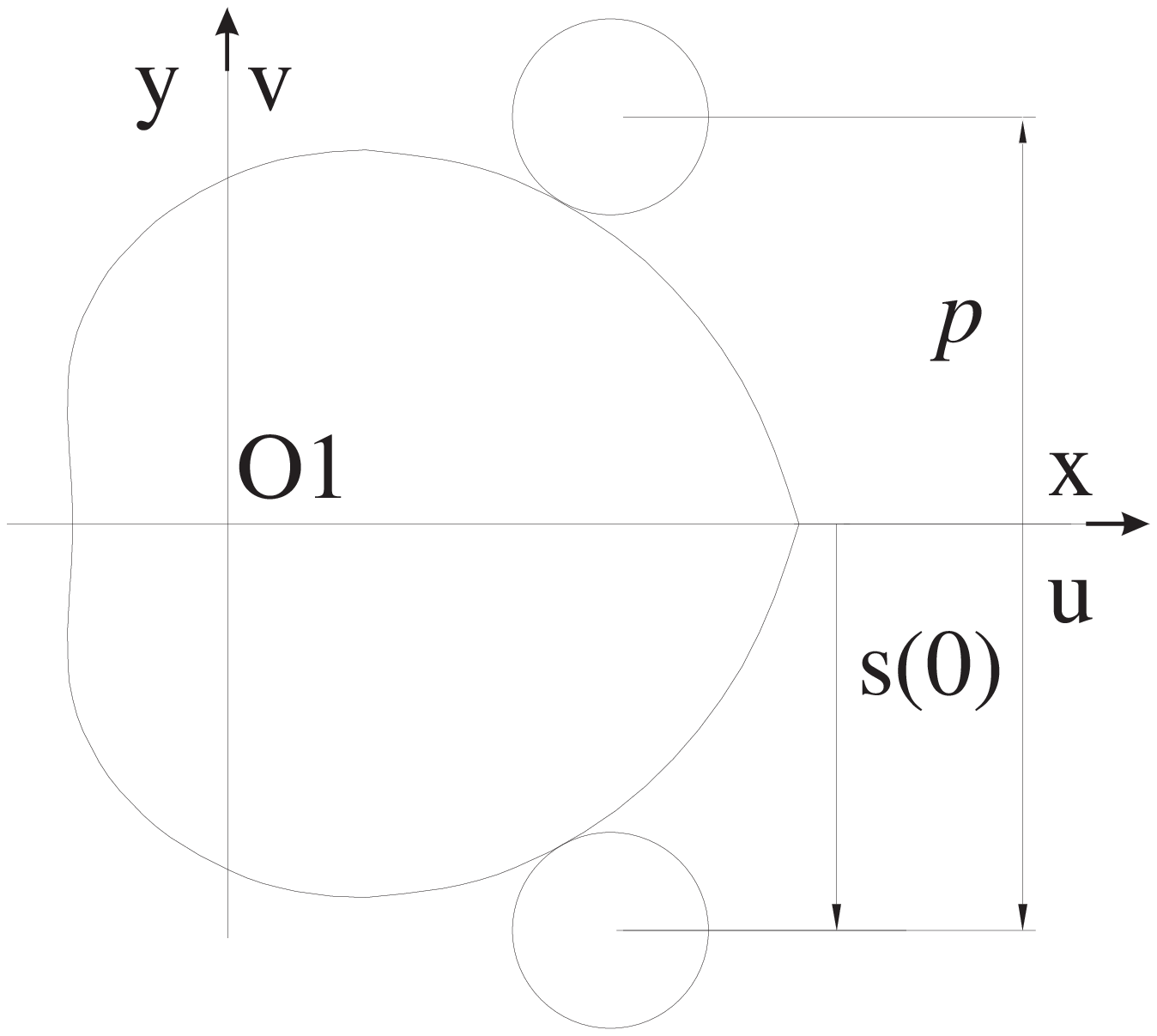,scale = 0.3}}
   \caption{Home configuration of the mechanism} \label{fig004}
 \end{center}
 \end{minipage}
 \end{figure}

Let the $x$-$y$ frame be fixed to the machine frame and the $u$-$v$
frame be attached to the cam, as depicted in Fig.~\ref{fig003}.
$O_{1}$ is the origin of both frames, $O_{2}$ is the center of the
roller, and $C$ is the contact point between the cam and the roller.
The geometric parameters are illustrated in the same figure. The
notation used in this figure is based on the general notation
introduced in \cite{Gonzalez-Palacios:1993}, namely, ($i$)~$p$ is
the pitch, {\it i.e.}, the distance between the center of two
rollers on the same side of the follower; ($ii$)~$e$ is the distance
between the axis of the cam and the line of centers of the rollers;
($iii$)~$a_{4}$ is the radius of the roller-bearing, {\it i.e.}, the
radius of the roller; ($iv$)~$\psi$ is the angle of rotation of the
cam, the input of the mechanism; ($v$)~$s$ is the position of the
center of the roller, {\it i.e}, the displacement of the follower,
which is the output of the mechanism; ($vi$)~$\mu$ is the pressure
angle; and ($vii$)~{\bf f} is the force transmitted from the cam to
the roller.

\begin{figure}[!ht]
 \begin{center}
     \psfrag{O1}{$O_1$}
     \psfrag{x}{$x$}
     \psfrag{y}{$y$}
     \psfrag{u}{$u$}
     \psfrag{v}{$v$}
     \psfrag{C}{$C$}
     \psfrag{Psi}{$\psi$}
     \subfigure[]{\epsfig{file =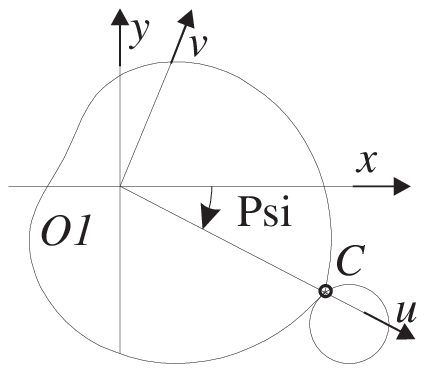,scale = 0.6}}
     \psfrag{x, u}{$x, u$}
     \psfrag{y, v}{$y, v$}
     \psfrag{delta}{$\Delta$}
     \psfrag{-u}{-$u$}
     \psfrag{-v}{-$v$}
     \subfigure[]{\epsfig{file =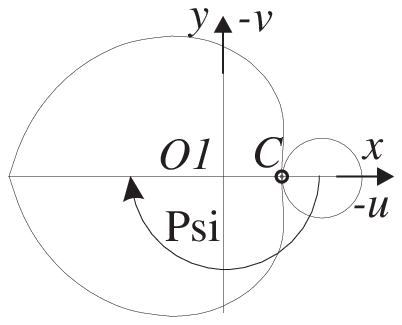,scale = 0.6}}
     \psfrag{u}{$u$}
     \psfrag{v}{$v$}
     \subfigure[]{~~~\epsfig{file =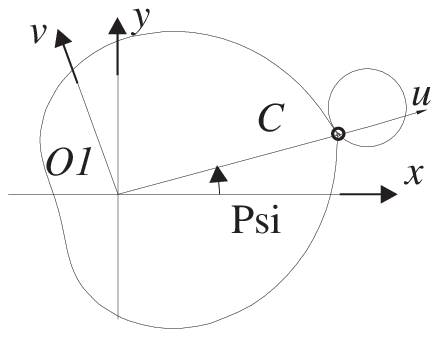,scale = 0.6}}
 \end{center}
 \caption{Orientations of the cam found when $v_c=0$: (a) $\psi=\Delta$;
  (b) $\psi=\pi$; and (c) $\psi=2\pi-\Delta$}
 \label{fig005}
 \end{figure}

The above parameters as well as the surface of contact on the cam,
are determined by the geometric relations derived from the
Aronhold-Kennedy Theorem~\cite{Waldron:1999}. As a matter of fact,
when the cam makes a complete turn ($\Delta\psi=2\pi$), the
displacement of the roller is equal to $p$, the distance between two
rollers on the same side of the roller-carrying slider ($\Delta s
=p$). Furthermore, if we consider that Fig.~\ref{fig004} illustrates
the home configuration of the roller, the latter is below the
$x$-axis when $\psi=0$. Therefore, $s(0)=-p/2$ and the input-output
function $s$ is defined as follows:
\begin{equation}
  s(\psi)=\frac{p}{2\pi}\psi-\frac{p}{2} \label{eq01}
 \end{equation}
The expressions of the first and second derivatives of $s(\psi)$ are
given by:
 \begin{eqnarray}
 s'(\psi)=p/(2\pi) \quad {\rm and} \quad s''(\psi)=0
 \label{eq02}
 \end{eqnarray}
The cam profile is determined by the displacement of the contact
point $C$ around the cam. The Cartesian coordinates of this point in
the $u$-$v$ frame take the form \cite{Gonzalez-Palacios:1993}
 \begin{subequations}
 \begin{eqnarray}
 u_{c}(\psi)\!\!\! &=&\!\!\! ~~b_{2} \cos \psi+(b_{3}-a_{4})\cos(\delta-\psi) \\
 v_{c}(\psi)\!\!\! &=&\!\!\! -b_{2} \sin \psi + (b_{3}-a_{4})\sin(\delta-\psi)
 \end{eqnarray}
 \label{eq04}
 \end{subequations}
\noindent the expression of coefficients $b_{2}$, $b_{3}$ and
$\delta$ being
 \begin{subequations}
 \begin{eqnarray}
 b_{2}  \!\!\!&=&\!\!\! -s'(\psi) \sin \alpha_{1} \\
 b_{3}  \!\!\!&=&\!\!\! \sqrt{(e+s'(\psi)\sin \alpha_{1})^{2}+(s(\psi)
 \sin\alpha_{1})^{2}} \\
 \delta \!\!\!&=&\!\!\! \arctan \left( \frac{-s(\psi) \sin\alpha_{1}}{e+s'(\psi) \sin \alpha_{1}} \right)
 \end{eqnarray}
 \label{eq05}
 \end{subequations}
\noindent where $\alpha_{1}$ is the directed angle between the axis
of the cam and the translating direction of the follower.
$\alpha_{1}$ is positive in the counterclockwise~(ccw) direction.
Considering the orientation adopted for the input angle $\psi$ and
the direction defined for the output $s$, as depicted in
Fig.~\ref{fig003},
 \begin{equation}
 \alpha_{1}=-\pi /2
 \label{eq06}
 \end{equation}
The nondimensional design parameter $\eta$ is defined below and will
be used extensively in what remains.
\begin{equation}
\eta= e/p \label{eq07}
\end{equation}
The expressions of $b_2$, $b_3$ and $\delta$ can be simplified using
Eqs.~(\ref{eq01}), (\ref{eq02}), (\ref{eq05}a--c), (\ref{eq06}) and
(\ref{eq07}):
 \begin{subequations}
 \begin{eqnarray}
 b_{2}  \!\!\!&=&\!\!\! \frac{p}{2\pi} \\
 b_{3}  \!\!\!&=&\!\!\! \frac{p}{2\pi}\sqrt{(2\pi \eta -1)^{2}+(\psi-\pi)^{2}} \\
 \delta \!\!\!&=&\!\!\! \arctan\left(\frac{\psi-\pi}{2\pi \eta -1} \right)
 \end{eqnarray}
 \label{eq08}
 \end{subequations}
From Eq.~(\ref{eq08}), $\eta$ cannot be equal to $1/(2\pi)$. That is
the first constraint on $\eta$. An {\it extended angle} $\Delta$ was
introduced in~\cite{Lee:2001} to know whether the cam profile can be
closed or not. Angle $\Delta$ is defined as a root of
Eq.~(\ref{eq04}). In the case of Slide-o-Cam, $\Delta$ is negative,
as shown in Fig.~\ref{fig005}. Consequently, the cam profile is
closed if and only if $\Delta \leq \psi \leq 2\pi-\Delta$.
\subsection{Pitch-Curve Determination}
The pitch curve is the trajectory of $O_{2}$, the center of the
roller, distinct from the trajectory of the contact point $C$, which
produces the cam profile. $(e,s)$ are the Cartesian coordinates of
point $O_{2}$ in the $x$-$y$ frame, as depicted in
Fig.~\ref{fig003}. Hence, the Cartesian coordinates of the
pitch-curve in the $u$-$v$ frame are
 \begin{subequations}
 \begin{eqnarray}
    u_{p}(\psi) \!\!\!& = \!\!\!& ~~e \cos \psi + s(\psi)\sin \psi
    \label{eq010}\\
    v_{p}(\psi) \!\!\!& = \!\!\!& -e  \sin \psi + s(\psi)\cos \psi
  \end{eqnarray}
 \end{subequations}
\subsection{Geometric Constraints on the Mechanism}
\begin{figure}[!ht]
\center
  \psfrag{O1}{$O_1$}
  \psfrag{x}{$x$}
  \psfrag{y}{$y$}
  \psfrag{p}{$p$}
  \psfrag{e}{$e$}
  \psfrag{b}{$b$}
  \psfrag{C}{$C$}
  \psfrag{a4}{$a_4$}
  \subfigure[]{\epsfig{file =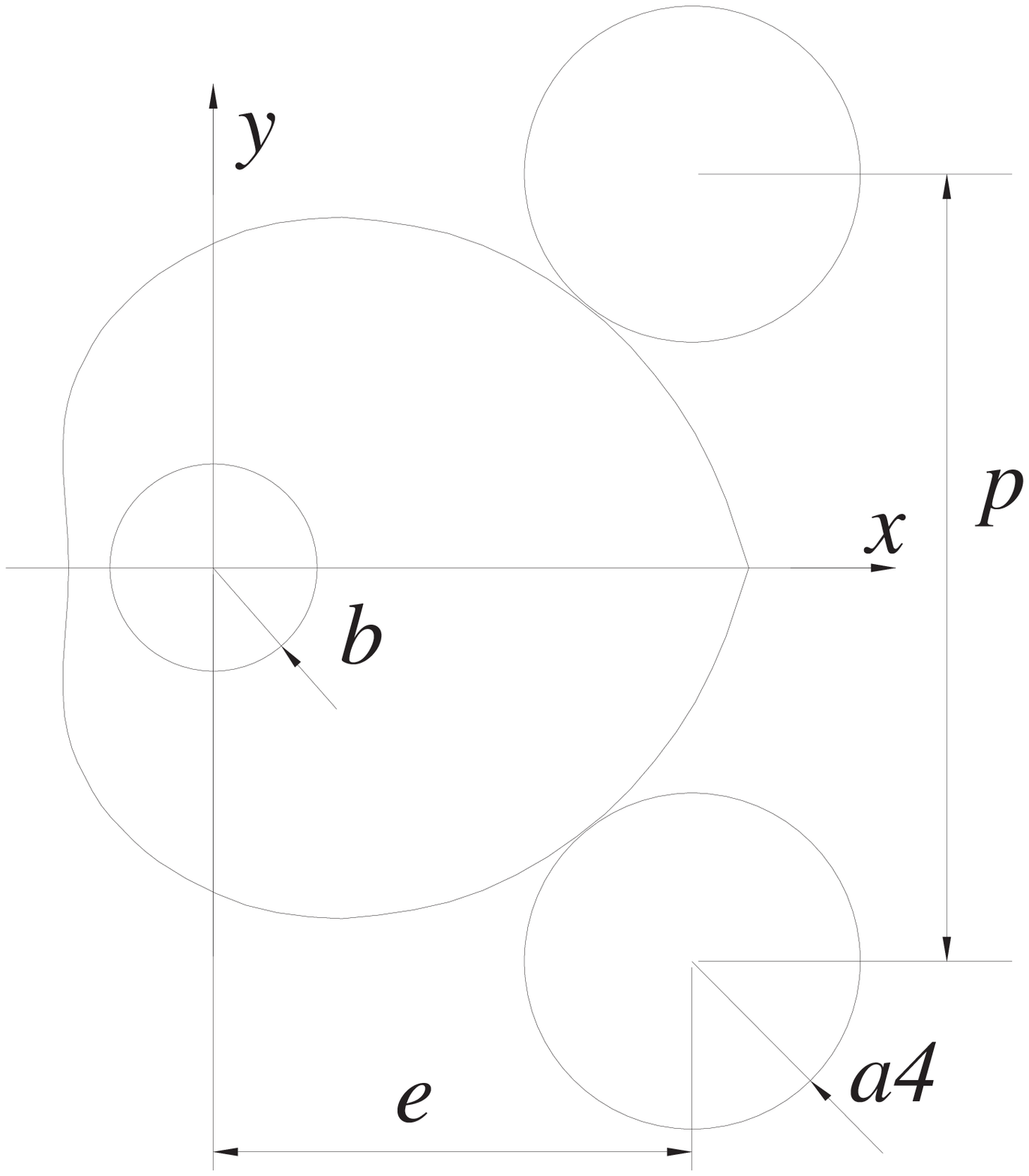,scale = 0.20}~~~~~~~}
  \subfigure[]{~~~~~~~~\epsfig{file =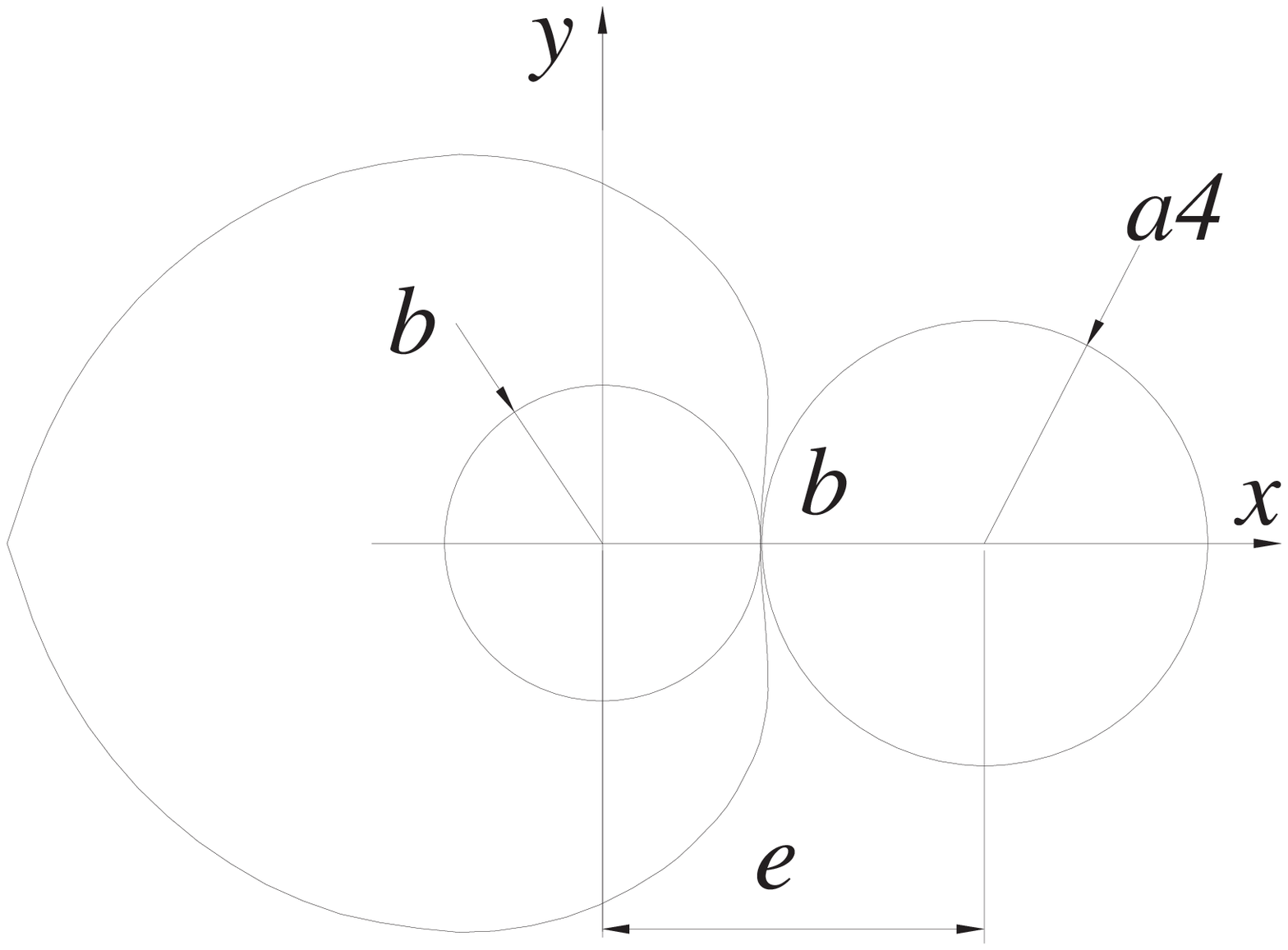,scale = 0.20}}
  \caption{Constraints on the radius of the roller: (a) $ a_{4} / p < 1/2$; and (b) $a_{4}/p \leq \eta -b/p$.}
  \label{fig006}
\end{figure}

In order to lead to a feasible mechanism, the radius $a_{4}$ of the
roller must satisfy two conditions, as shown in Fig.~\ref{fig006}a:
\begin{itemize}
\item  Two consecutive rollers on the same side of the roller-carrying
slider can not collide. Since $p$ is the distance between the
centers of two consecutive rollers, the constraint $2a_{4} < p$ has
to be respected.
\item  Likewise, the radius $b$ of the camshaft has to be considered.
Therefore, the following condition has to be respected: $a_4 + b
\leq e$, which can written in terms of $\eta$:
\end{itemize}
     \begin{equation}
     \label{eq012}
     a_{4} / p \leq \eta -b / p
     \end{equation}
According to the initial configuration of the roller, as depicted in
Fig.~\ref{fig004}, the $v$-component of the Cartesian coordinate of
contact point $C$ is negative in this configuration, {\it i.e.}, $
v_{c}(0) \leq 0 $. Moreover, from the expression of $v_{c}(\psi)$
and for parameters $b_{3}$ and $\delta$ given in Eqs.~(\ref{eq04}b),
(\ref{eq08}b \& c), respectively, the above relation leads to the
condition:
 \beqa
  \left( \frac{p}{2\pi a_4}\sqrt{(2\pi \eta-1)^{2}+(-\pi)^{2}} - 1 \right)\nonumber  \\
  \sin \left[\arctan\left(\frac{-\pi}{2\pi \eta -1}\right) \right]
  \leq 0 \nonumber
  \eeqa
It turns out that the constraint $v_{c}(0) \leq 0$ leads to a constraint on $\eta$,
namely \cite{Renotte:2004},
\begin{equation}
 \label{eq013}
 \eta> 1/(2\pi)
\end{equation}
\subsection{Curvature of the Cam Profile}
The curvature of any planar parametric curve, in terms of the
Cartesian coordinates $u$ and $v$, and parameterized with any
parameter $\psi$, is given by \cite{Angeles:1991}:
 \begin{equation}
 \label{eq1}
 \kappa=\frac{v'(\psi)u''(\psi)-u'(\psi)v''(\psi)}{[u'(\psi)^{2}+v'(\psi)^{2}]^{3/2}}
 \end{equation}
The curvature $\kappa_{p}$ of the pitch curve is given in \cite{Renotte:2004} as
 \begin{equation}
 \label{eq6}
 \kappa_{p}=\frac{2\pi}{p} \frac{[(\psi-\pi)^{2}+2(2\pi
\eta-1)(\pi \eta-1)]}{[(\psi-\pi)^{2}+(2\pi \eta-1)^{2}]^{3/2}}
 \end{equation}
provided that the denominator never vanishes for any value of $\psi$, {\it i.e.}, provided that
 \begin{equation}
 \label{eq7}
 \eta \neq 1/(2\pi)
 \end{equation}
\noindent Let $\rho_{c}$ and $\rho_{p}$ be the radii of curvature of both the cam profile and the pitch curve, respectively, and $\kappa_{c}$ the curvature of the cam profile. Since the curvature is the reciprocal of the radius of curvature, we have $\rho_{c} = 1/\kappa_{c}$ and $\rho_{p} = 1/\kappa_{p}$. Furthermore, due to
the definition of the pitch curve, it is apparent that 
 \begin{equation}  
   \label{eq8} \rho_{p} = \rho_{c} + a_{4}
 \end{equation}
Writing Eq.~(\ref{eq8}) in terms of $\kappa_{c}$ and $\kappa_{p}$, we obtain the curvature of the cam profile as
 \begin{equation}
 \kappa_{c}=\frac{\kappa_{p}}{1-a_{4} \kappa_{p}}
 \end{equation}
\subsection{Physical constraints}
Let us assume that the surfaces of contact are ideal, smooth and
dry, with negligible friction. The relations between the forces of
contact are described below:
\begin{itemize}
\item [-] Two relations follow from the strength of materials.
Besides, the bearing shafts are subject to shearing, whereas the
camshafts are subject to shearing and bending. Consequently, we come
up with the following relations:
\end{itemize}
\begin{eqnarray}
    \tau c_{max} &\geq& 8 M_t
    \left(\frac{2}{\pi \, \phi_{cam}^3}+\frac{1}{p \, \phi_{cam}^2}\right) \label{constrainstressa} \\
    \tau b_{max} &\geq& \frac{8 M_t}{p \, \phi_{bear}^2} \label{constrainstressb}
\end{eqnarray}
where,
\begin{description}
  \item [$\phi_{cam}$] is the diameter of the camshaft ($\phi_{cam}= 2(e - a_4)$);
  \item [$\phi_{bear}$] is the diameter of the bearing's shaft ($\phi_{bear}= 2a_4$);
  \item [$M_t$] is the torque applied to the camshaft;
  \item [$\tau c_{max}$] is the maximum allowable stress inside the cam axis which cannot be exceeded in the camshaft;
  \item [$\tau b_{max}$] is the maximum stress inside the bearing's axis which cannot be exceeded in the bearing shaft;
\end{description}

Let us assume that the material of the cam and the one of the roller
have the same Young modulus $E$ and Poisson ratio. The Hertz's
formula~\cite{Golovin:2005} yields the maximum pressure of contact
between the cams and rollers. Depending on the material, surface
roughness and possible surface treatment, the Hertz pressure $P_{Hertz}$ must remain
smaller than a maximum value $P_{max}$.
\begin{equation}\label{constrainhertz}
        P_{Hertz} \equiv 0.418 \sqrt{\frac{F \, E}{a \, r_{eq}}} \leq P_{max}
\end{equation}
where,
\begin{description}
  \item[$F$] is the axial load, 
  $\displaystyle{F = \sqrt{\left(M_t \, \frac{2 \pi}{p}\right)^2+\left(M_t \, \frac{2 \pi}{p \tan(\delta)}\right)^2}}$,
  \item[$r_{eq}$] is the equivalent radius of the contact,
  \item[$a$] is the width of the cam and the roller, 
  \[\displaystyle{r_{eq} = 1 / \left(\frac{2}{\phi_{cam}}+ \frac{2}{\phi_{bear}}\right)}\]
  \item [$P_{max}$] is the maximum pressure acceptable between two surfaces of contact.
\end{description}

\begin{figure}[!htbp]
  \psfrag{F}{$\negr F$}
  \psfrag{R1}{$\negr r_1$}
  \psfrag{R2}{$\negr r_2$}
  \psfrag{(a)}{(a)}
  \psfrag{(b)}{(b)}
  \centerline{\epsfig{file = 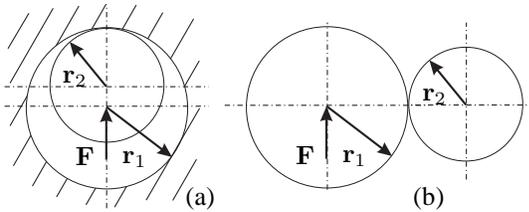,scale = 0.4}}
  \caption{Equivalent radius of contact with (a) a positive and a negative radius and (b) two positive radius}
  \label{fig:rayoneq}
\end{figure}
To minimize the Hertz pressure, we must maximize the equivalent
radius of contact. Figure~\ref{fig:rayoneq}(a) shows an optimal
configuration to minimize the Hertz pressure \cite{Carra:2004}, i.e.
the stresses bear opposite sign. The roller of radius $r_1$ is
included in the roller of radius $r_2$ being smaller than $r_1$ but
close to it. However, this layout does not occur in Slide-o-Cam
because the contact changes between the cam and rollers. Indeed,
Fig.~\ref{fig:rayoneq}(b) depicts the actual configuration. In this
case, the only way to minimize the Hertz pressure is to maximize the
diameter of the roller and the curvature of the cam.

The Hertz pressure is evaluated only when the cam pushes the roller. The active interval is \cite{Renotte:2004}:
\[
 \frac{\pi}{n}-\Delta \leq \psi \leq
 \frac{2\pi}{n}-\Delta
 \]
\subsection{Implementation}
\begin{figure}[htb]
 \begin{center}
   \centerline{\epsfig{file = 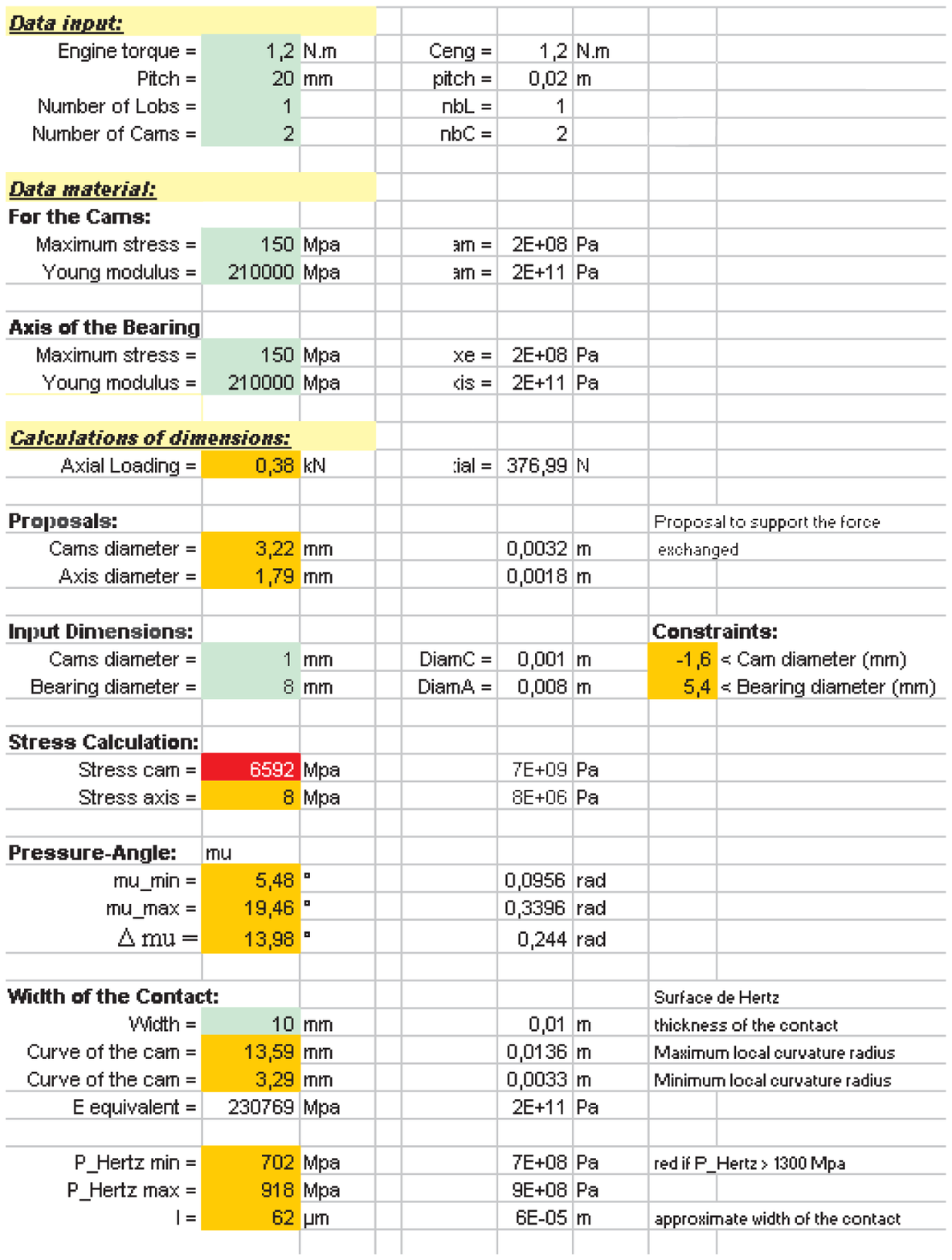,scale = 0.6}}
   \caption{Worksheet of Slide-o-Cam}
  \label{fig0018}
 \end{center}
\end{figure}

A graphic user interface (GUI) based on the synthesis of planar cam
mechanism is implemented in Excel, as shown in Fig.~\ref{fig0018}.
This GUI allows the user to determine the dimensions of the cams. The
value in blue cells have to be defined by the user. The results are
shown in the yellow cells whereas the critical values are displayed
in red cells.
The GUI allows the user to test and visualize the cam profile
generated.
\begin{figure}[htb]
 \begin{center}
   \psfrag{a}{\small $a$}
   \psfrag{formula}{\small formula}
   \psfrag{Input}{\small Input}
   \psfrag{P=P0}{\small $P=P_0$}
   \psfrag{Cam}{\small Cam}
   \psfrag{profile}{\small profile}
   \psfrag{Load}{\small Load}
   \psfrag{Dbearing}{\small $\phi_{bearing}$}
   \psfrag{Dcam}{\small $\phi_{cam}$}
   \psfrag{Hertz's}{\small Hertz's}
   \psfrag{Impossible}{\small Impossible}
   \psfrag{Yes}{\small Yes}
   \psfrag{nl=1}{\small $n_l=1$}
   \psfrag{nc=2}{\small $n_c=2$}
   \psfrag{Deltamu}{\small $\Delta_{\mu}$}
   \psfrag{Mumax}{\small $\mu_{max}$}
   \psfrag{No}{\small No}
   \psfrag{Ok?}{\small Ok?}
   \psfrag{Stress}{\small Stress}
   \psfrag{calculation}{\small calculation}
   \psfrag{P=P+P0}{{\small $P=P+P_O$}}
   \psfrag{nc=nc+1}{\small $n_c=n_c+1$}
   \centerline{\epsfig{file = 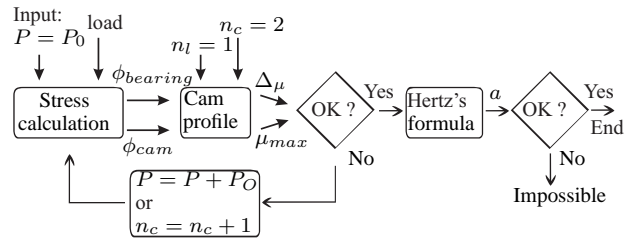,scale = 0.65}}
   \caption{Scheme of the design process}
  \label{fig0020}
 \end{center}
\end{figure}
The algorithm underlying this GUI is shown in Fig.~\ref{fig0020}.
\begin{enumerate}
  \item The engine torque and the pitch $p$ are assigned by the user;
  \item for a given material of the camshaft and the roller, the minimum diameters are
computed;
  \item the shape of the cam and the pressure angle are defined;
  \item in case the pressure angle is too high, the initial parameters and the number of cams are
reassigned;
  \item in the other case, the Hertz pressure is evaluated to define the
width of the cam and roller.
\end{enumerate}

%% file: 04_DesignParamPressAngleHertzPress.tex
\section{Influence of the design parameters on the pressure angle and the Hertz pressure}
\subsection{Analysis}
The maximal value of the Hertz pressure depends on several parameters, namely, the number of conjugate cams, the material of the parts in contact, the geometry of the cams, and the load applied. The pressure angle can be optimized with regards to the number of cams and the shape of the cams~\cite{Renotte:2004}. There are many ways to minimize the Hertz pressure:
\begin{itemize}
  \item increase the number of conjugate cams in order to decrease the maximum pressure angle. Nevertheless, the number of simultaneous contact lines does not increase.
  \item decrease the axial load. It can be done by increasing the pitch of the transmission.
  \item choose a material with a lower Young modulus. However, when the material is more compliant, the maximum pressure acceptable decreases because the plastic domain occurs for smaller stress.
  \item decrease the minimum radius of the cam. This feature will be used in the next section.
  \item increase the width of the cam and the roller.
\end{itemize}
\subsection{Design strategies}
\begin{figure}[htb]
 \begin{center}
   \psfrag{(a)}{\small (a)}
   \psfrag{(b)}{\small (b)}
   \psfrag{Cam}{Cam}
   \psfrag{Roller}{Roller}
   \psfrag{Shaft}{Shaft}
   \psfrag{x}{$x$}
   \psfrag{y}{$y$}
   \centerline{\epsfig{file = 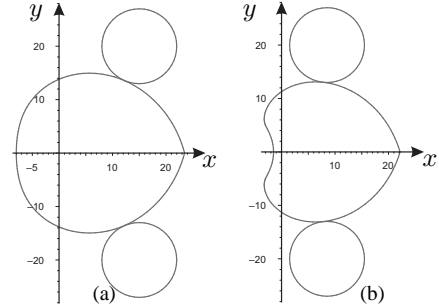,scale = 0.7}}
  \caption{Design strategies of Slide-o-Cam with
  (a) $\phi_{cam}= 16mm$, $a_4= 7 mm$, $p=40 mm$, $\mu_{max}= 53.8^{\circ}$ and $\Delta \mu = 35.6 ^{\circ}$ and
  (b) $\phi_{cam}= 3mm$, $a_4= 7 mm$, $p=40 mm$, $\mu_{max}= 15.6^{\circ}$ and $\Delta \mu = 11.1^{\circ}$}
  \label{fig0010}
 \end{center}
 \end{figure}

\begin{figure}[htb]
 \begin{center}
   \psfrag{Box with two cams}{\small {Box with two cams}}
   \psfrag{(a)}{\small (a)}
   \psfrag{(b)}{\small (b)}
   \centerline{\epsfig{file = 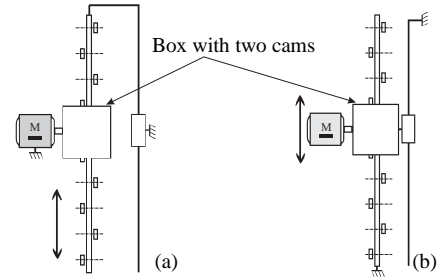,scale = 0.45}}
   \caption{Parametrization of Slide-o-Cam with (a) motor fixed and (b) motor fixed on the cam mechanism}
  \label{fig0011}
 \end{center}
 \end{figure}

In this section, we present four strategies to minimize the radius
of the cam:
\begin{enumerate}
\item The lower $e$, the distance between the cam axis and the bearing axis, the lower the maximum pressure angle $\mu_{max}$ and its range $\Delta \mu$. This involves the removal of the bearing between the shaft and the cam and its offset on the base.
\item With the first design strategy, the bearings can collide with the camshaft. In order to figure out this issue, the distance $e$ is increased. However, the advantages of the first solution are not maintained. In this case, a good compromise is to use two lines of followers on both sides of the cams.
\item Another solution is to use cams assembled on the shaft. This design is more complex, but has some advantages. For instance, by means of a FEM, the optimal diameters of the camshaft and the roller are equal to 16~mm and 14~mm, respectively. For these values, $\mu_{max}= 53.8^{\circ}$ and $\Delta \mu = 35.6 ^{\circ}$ as illustrated in~Fig.~\ref{fig0010}(a). With an inserted cam, we come up with a better design: $\mu_{max}= 15.6^{\circ}$, $\Delta \mu = 11.2^{\circ}$ and $\phi_{cam}=$~3~mm as illustrated in Fig.~\ref{fig0010}(b).
\item In the first design of the Slide-o-Cam transmission, the motor is fixed, as shown in Fig.~\ref{fig0011}(a). We have the following drawbacks:
\begin{itemize}
  \item For the same length of displacement, the length of the module is double because of guidance;
  \item The effector being positioned  at the end of the follower, the latter is subjected to bending;
  \item Many parts are assembled between the motor and the effector. Consequently, the stiffness of the unit is smaller.
\end{itemize}
To figure out with these issues, the motor can be linked to the
end-effector so that they can move altogether, as shown in~Fig.~\ref{fig0011}(b).
\end{enumerate}

\begin{figure}[htb]
 \begin{center}
   \centerline{\epsfig{file = 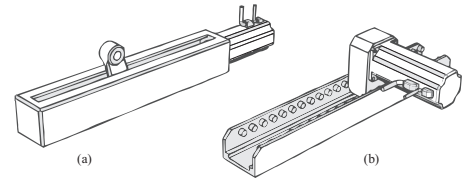,scale = 0.5}}
  \caption{Conversion of motion from rotational to translational made by (a) a ball-screws and (b) Slide-o-Cam with motor embedded}
  \label{fig0014}
 \end{center}
 \end{figure}

If we apply the previous design strategies, we will replace the classical ball-screws, Fig.~\ref{fig0014}(a), by a new transmission where the motor moves with the camshaft and the effector is attached to it, Fig.~\ref{fig0014}(b).

%% file: 05_Applications.tex
\section{Implementation of the transmissions to the Orthoglide}
\begin{figure}[htb]
 \begin{center}
    \epsfig{file = 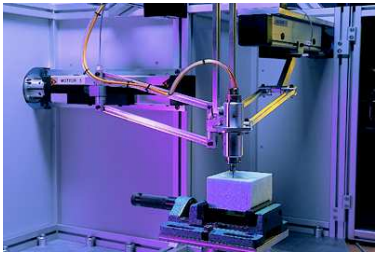,scale = 0.35}
    \caption{The Orthoglide (\copyright CNRS Phototh\`eque /  CARLSON Leif)}
    \label{fig002}
 \end{center}
\end{figure}

A motivation of this research work is to design a \textit{Slide-o-Cam} transmission for high-speed machines. As mentioned in the introduction, this mechanism should be suitable for the \textit{Orthoglide}, which is a low power machine tool, as shown in Fig.~\ref{fig002} \cite{Chablat:2003}. Here is a list of its features:
\begin{itemize}
    \item [-] ball screw engine torque $= 1.2$ N.m;
    \item [-] ball screw engine velocity $= 0$ to $3000$~rpm;
    \item [-] ball screw pitch $= 20$ mm/turn;
    \item [-] axial static load $= 376$ N;
    \item [-] stiffness $= 130$ N/$\mu$m.
\end{itemize}

Let us assume that the maximum stress that the shafts can support is equal to $150$~MPa. We can compute the minimum diameter of the bearing shaft $\phi_{bear}$ and the cam shaft $\phi_{cam}$ to transmit the load. The minimum diameter of the bearing shaft $\phi_{bear}$ is equal to 1.8~mm. Likewise, the minimum diameter of the camshaft $\phi_{cam}$ is equal to 3.75~mm. 

\begin{figure}
 \begin{center}
   \psfrag{-10}{\tiny -10}
   \psfrag{-5}{\tiny -5}
   \psfrag{-2}{\tiny -2}
   \psfrag{2}{\tiny 2}
   \psfrag{4}{\tiny 4}
   \psfrag{5}{\tiny 5}
   \psfrag{6}{\tiny 6}
   \psfrag{8}{\tiny 8}
   \psfrag{10}{\tiny 10}
   \psfrag{12}{\tiny 12}
   \psfrag{(a)}{(a)}
   \psfrag{(b)}{(b)}
   \psfrag{(c)}{(c)}
   \psfrag{(c)}{(d)}
   \psfrag{Cam}{Cam}
   \psfrag{Roller}{Roller}
   \psfrag{Shaft}{Shaft}
   \psfrag{x}{$x$}
   \psfrag{y}{$y$}
   \centerline{\epsfig{file = 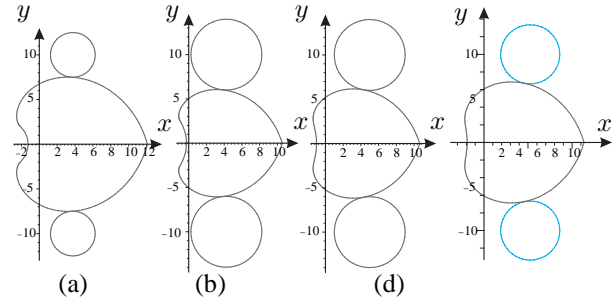,scale = 0.65}}
  \caption{Design strategies of Slide-o-Cam for the Orthoglide with
  (a)  $\phi_{cam}=2.5 mm$, $\phi_{bear}= 5 mm$,    $\mu_{max} = 8.0^{\circ}$
and $\Delta {\mu}=5.7^{\circ}$,
  (b)  $\phi_{cam}=0.5 mm$, $\phi_{bear}= 8 mm$,    $\mu_{max} = 15.8^{\circ}$
and $\Delta {\mu}=11.3^{\circ}$,
  (c) $\phi_{cam}=2 mm$, $\phi_{bear}= 8mm$,    $\mu_{max} = 26.6^{\circ}$ and
$\Delta {\mu}=19.0^{\circ}$
   and
  (d) $\phi_{cam}=3.8mm$, $\phi_{bear}= 6.7mm$,  $\mu_{max} = 30.0^{\circ}$
and $\Delta {\mu}=21.5^{\circ}$}
  \label{fig0012}
 \end{center}
 \end{figure}

Table~\ref{table1} and Fig.~\ref{fig0012} depict four design strategies applied to the Orthoglide. In case of high-speed
operations, angular velocities of cams are higher than 50~rpm. Therefore, the pressure angle has to be smaller than 30$^{\circ}$. For instance, with case study~(d) the Hertz pressure is a minimum but the pressure angle is a maximum.

From cases (a) to (d), the local radius of the cam increases from 0.08~mm to 1.6~mm. This feature verifies the assumption of section~II.D related to the minimization of the Hertz pressure. For the optimal design,  the Orthoglide can transmit the load with $\phi_{cam}$~=~3.8~mm.
\begin{table}
  \begin{center}
    \begin{tabular}{|c|c|c|c|c|}
    \hline
    Case \# & $P_{max}$ & $P_{min}$ & $\mu_{max}$ & $r_{cam_{min}}$\\
    \hline
    (a) & 786 MPa & 579 MPa   & $8.0^{\circ}$ & 1.13mm\\ 
    (b) & 933 MPa & 492 MPa   & $15.8^{\circ}$ & 0.08mm\\ 
    (c) & 732 MPa  & 492 MPa  & $26.6^{\circ}$ & 1.44mm\\ 
    (d) & 689 MPa & 522 MPa & $30.0^{\circ}$ & 1.6mm\\ 
    \hline
    \end{tabular}
    \caption{The Hertz pressure obtained for four design strategies with $a=20mm$}
    \label{table1}
  \end{center}
\end{table}

%% file: 06_Conclusions.tex
\section{Acknowledgments}
This research work was made in the context of an exchange between McGill University and Centrale Nantes, which allows Lucas Chabert to work six months in Montreal. Moreover, the authors would like to thank Prof. Jorge Angeles for his great help and good advice.
\section{Conclusions}
New design strategies were presented in this paper to minimize
the Hertz pressure in the rollers and the cam of the Slide-o-Cam
mechanism. A graphic user interface was developed to synthesize
planar cam mechanisms, considering their physical constraints. The
cam profiles generated can be used to realize CAD models.

The main contribution of this research work lies in the study of the influence of
the pressure angle on the Hertz pressure. As a matter of fact, the
smaller the pressure angle of a cam, the higher its Hertz pressure.
In order to reduce the maximum Hertz pressure of a cam, we defined
its parameters in order to keep the pressure angle smaller than~$30^{\circ}$.